\documentclass[letterpaper, 10 pt, conference]{ieeeconf}
\IEEEoverridecommandlockouts
\usepackage{cite}
\usepackage{amsmath,amssymb,amsfonts}
\usepackage{algorithmic}
\usepackage{graphicx}
\usepackage{textcomp}
\usepackage{url}
\usepackage[table]{xcolor}
\def\BibTeX{{\rm B\kern-.05em{\sc i\kern-.025em b}\kern-.08em
    T\kern-.1667em\lower.7ex\hbox{E}\kern-.125emX}}
    
\begin{document}

\title{{\LARGE \bf
From Language Priors to Field Adaptation:\\Preference Learning for Traversability Estimation*
}
\thanks{*This work was partly supported by the city of Vienna (MA23 – Economic Affairs, Labour and Statistics) through the project Stadt Wien Kompetenzteam für Drohnentechnik in der Fachhochschulausbildung (DrohnFH, MA23 project 35-02).}
}

\author{Simon Schwaiger$^{1,2}$, David Seyser$^{2}$, Alessandro Scherl$^{2,3}$\\Zlatan Ajanovi{\'c}$^{2}$, Wilfried Wöber$^{4}$, Gerald Steinbauer-Wagner$^{1}$
\thanks{$^{1}$Graz University of Technology, Institute of Software Engineering and Artificial Intelligence, Graz, Austria}%
\thanks{$^{2}$University of Applied Sciences Technikum Wien, Research Group Automation Technology and Robotic Systems, Vienna, Austria {\tt\small schwaige@technikum-wien.at}}%
\thanks{$^{3}$University of Alicante, Department of Computer Technology, Alicante, Spain}%
\thanks{$^{4}$University of Natural Resources and Life Sciences, Institute for Integrative Nature Conservation Research, Vienna, Austria}%
}

\newcommand{\firstbest}[1]{\cellcolor{green!25}\textbf{#1}}
\newcommand{\secondbest}[1]{\cellcolor{yellow!30}\underline{#1}}
\newcommand{\thirdbest}[1]{\cellcolor{orange!30}#1}

\maketitle

\begin{abstract}
Image-based traversability estimation is inherently dependent on the robot platform, deployment domain, and mission preferences, which limits the applicability of purpose-trained models.
To facilitate domain adaptation, this work aims to reduce the number of required annotations in the target domain using sample-efficient preference learning.
Our method represents traversability through von Mises-Fisher mixture prototypes in a frozen vision-language feature space.
Relative natural-language rules provide a commonsense prior, while sparse relative image annotations adapt the prototype directions and utilities to a target domain through computationally and sample-efficient fine-tuning.
Experiments on WayFAST demonstrate accuracy competitive with end-to-end trained estimators while enabling sample-efficient image-based adaptation.
Qualitative experiments further demonstrate the language prior's zero shot applicability and the fine-tuned estimator's improved dense prediction on semantic maps.
Evaluation is complemented via semantic interpretation of learned prototypes by dissecting semantically close natural language prompts.
Code and trained estimators available at \url{https://resireg.github.io}.
\end{abstract}

\section{Introduction}

Image-based traversability estimation (TE) is important for autonomous outdoor navigation since it incorporates semantic differences between surfaces that might be geometrically similar but have different traversability properties.
However, TE is challenging due to the lack of platform- and domain-agnostic ground truth data.
Furthermore, the same terrain may have different traversability costs for different robots and application domains \cite{eder2023traversability,capezzuto2026towards,germann2025towards}.
Existing methods have tackled TE by i) learning physical capability from robot-specific field experience \cite{frey2024roadrunner,frey2023fast,gasparino2024wayfaster,roth2025learned,ruetz2025online,jung2024vstrong}, ii) learning from absolute or relative human supervision \cite{ruetz2024foresttrav,schreiber2024wrizz,schreiber2025do,mao2025pacer,karnan2024wait,mao2025terrain}, and iii) using language and foundation models as broad priors \cite{gummadi2025zest,potnis2026catnav,rana2026overseec,capezzuto2026towards}.
However, these methods either require i) physical robot experience, limiting zero-shot domain-adaptation, ii) platform-specific annotations, or iii) expensive computational power to provide semantically broad predictions, not suitable for targeted domain adaptation from experience. 

\begin{figure}[t]
    \centering
    \includegraphics[width=0.99\linewidth]{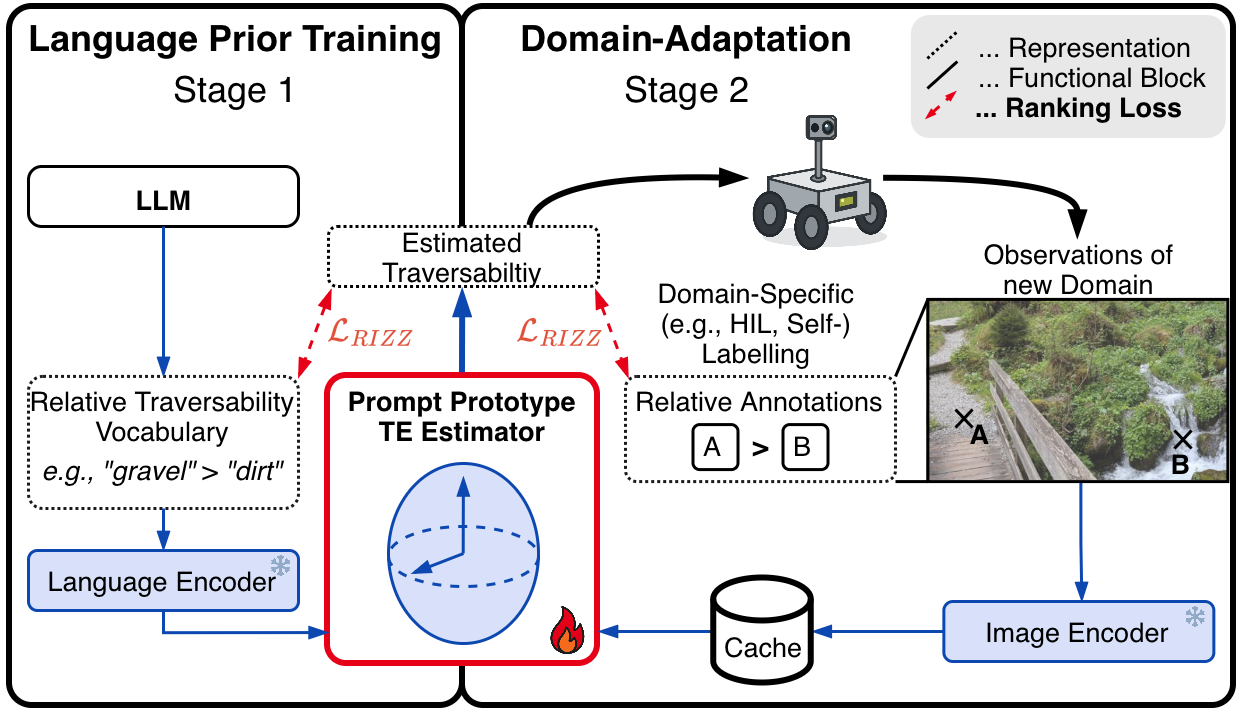}
    \caption{\textbf{Language-Initialised Traversability Adaptation.} In Stage~1, an LLM-generated vocabulary of relative traversability rules is embedded using a frozen language encoder to initialise the prompt prototypes. In Stage~2, the estimator is adapted to a new deployment domain using relative image annotations. This provides an initial commonsense traversability estimate while enabling lightweight and sample-efficient adaptation. Snowflake indicates a frozen VLM's encoders.}
    \label{fig:teaser_fig}
    \vspace{-2mm}
\end{figure}

We hypothesise that traversability does not need to be relearned from scratch as a dense visual prediction problem for each domain transfer.
We leverage terrain structure encoded in a frozen vision language model (VLM) \cite{schwaiger2026resireg}.
The VLM feature space, aligned across frozen image and language encoders, is reused across application domains.
This allows our method to be used zero-shot on out-of-distribution domains using a generic language prior, while enabling subsequent fine-tuning on image-based field data.
Instead of training a prediction head on top of a frozen foundation model, our method models traversability using von Mises-Fisher (vMF) mixture \cite{banerjee2005clustering} responsibilities over a set of prototypes with learned utility in vision-language feature space.
These prototypes are learned from relative annotations, both in language and vision, allowing a platform-agnostic language prior to bootstrap image-based fine-tuning. 

Our method (see Fig. \ref{fig:teaser_fig}) is competitive in terms of prediction accuracy with end-to-end trained models \cite{schreiber2024wrizz}, given the same amount of annotations, while offering language-based initialisation and sample-efficient learning from human annotations. 
To this end, we combine the principle of visual prototypes in robotics \cite{schwaiger2025otas} with recent VLM-based architectures \cite{schwaiger2026resireg}, and extend this architecture with alternative mixture modes to extend the state of the art and outperform alternative approaches.
To our knowledge, this combination of language-based prior and sample-efficient adaptation has not previously been demonstrated for traversability estimation.
The sole reliance on relative labels also allows us to incorporate new information sources, such as human-in-the-loop, self-labelling, or large language model (LLM) and large vision language model (LVLM)-generated relative image labels \cite{peng2025pigeon}.
Our experiments evaluate zero-shot and data-efficient adaptation, competitive fine-tuning accuracy, as well as downstream language-alignment of learned prototypes.

\section{Background}

For mobile robots, TE provides terrain-dependent cost information for path and motion planning.
Estimated traversability cost can differ for surfaces that are of identical geometry, for different robotic embodiments, and across application domains with different risk tolerances \cite{eder2023traversability,capezzuto2026towards,germann2025towards}. 
Traversability is thus a function of terrain, platform, and mission, and no single label can be correct for every robot or every task. 
Therefore, even though multiple regimes have been established to collect traversability data, no TE training data source can be used as ground truth across domains.

\textbf{Traversability Annotations.}
Three main annotation types for TE have been established in literature. 
First, self-supervised labels are generated through robot experience, either as proprioceptive signals such as velocity-tracking error, torque, slip, and collisions \cite{gasparino2024wayfaster,frey2024roadrunner,ruetz2024foresttrav,roth2025learned,ruetz2025online} or as hindsight labels that mark successfully traversed terrain as traversable and leave the rest unlabeled \cite{frey2023fast,jung2024vstrong}.
Secondly, an alternative source is human-generated labels, either as absolute labels (classes, scores) \cite{ruetz2024foresttrav,karnan2024wait,mao2025pacer,mao2025terrain} or as relative labels for different image regions \cite{schreiber2024wrizz,schreiber2025do}.
Finally, the most recent option is to generate labels using large language models (LLMs), which either generate cost labels \cite{gummadi2025zest,potnis2026catnav,rana2026overseec} or select navigation goals \cite{peng2025pigeon}.

Within these regimes, relative annotations have the advantage of being retrievable from absolute supervision signals.
Furthermore, the scale of absolute labels is tied to the platform they were estimated on, whereas ordering is invariant to the arbitrary numerical scaling of traversability. 
Therefore, relative labels offer increased flexibility for downstream estimator training w.r.t. training data for supervision.
However, this flexibility comes at the cost of information: an ordering says which region is easier, not by how much, which makes it harder to learn from, especially if the annotations are sparse.
Existing TE methods balance this trade-off either with end-to-end training, which reaches high accuracy at the cost of many labels \cite{schreiber2024wrizz}, or with few-shot adaptation on frozen features, which needs few labels but reaches lower accuracy \cite{schreiber2025do}.
This accuracy-adaptability trade-off is further complicated by both regimes' inability to initialise from generic open-world knowledge, e.g., provided by LLMs.
This is detrimental to open-world robots exploring environments whose conditions constantly change.
These robots simultaneously require short onboarding time as well as the ability to adapt to domain-specific TE preferences.

\textbf{Preference Learning.}
Preference learning \cite{celemin2022interactive}, whose goal is to learn from relative feedback in evaluative space, uniquely fits TE due to the absence of objective traversability scale.
A scalar score $s(x)$ is assigned to each item, and pairs are supervised only through their difference $\Delta = s(x_B) - s(x_A)$ with the goal of matching teacher preferences.
This is consistent with existing relative supervision in TE, where a ranking loss pushes $\Delta$ past a margin in the labelled direction and toward zero for ties \cite{schreiber2024wrizz}.
We therefore introduce a preference learning method that builds upon this established training objective, while, contrary to the state of the art, simultaneously enabling initialisation from a commonsense natural language prior and sample-efficient fine-tuning to adapt to new domains.


\section{Method}

\begin{figure*}[!ht]
    \vspace{2mm} 
    \centering
    \includegraphics[width=0.78\linewidth]{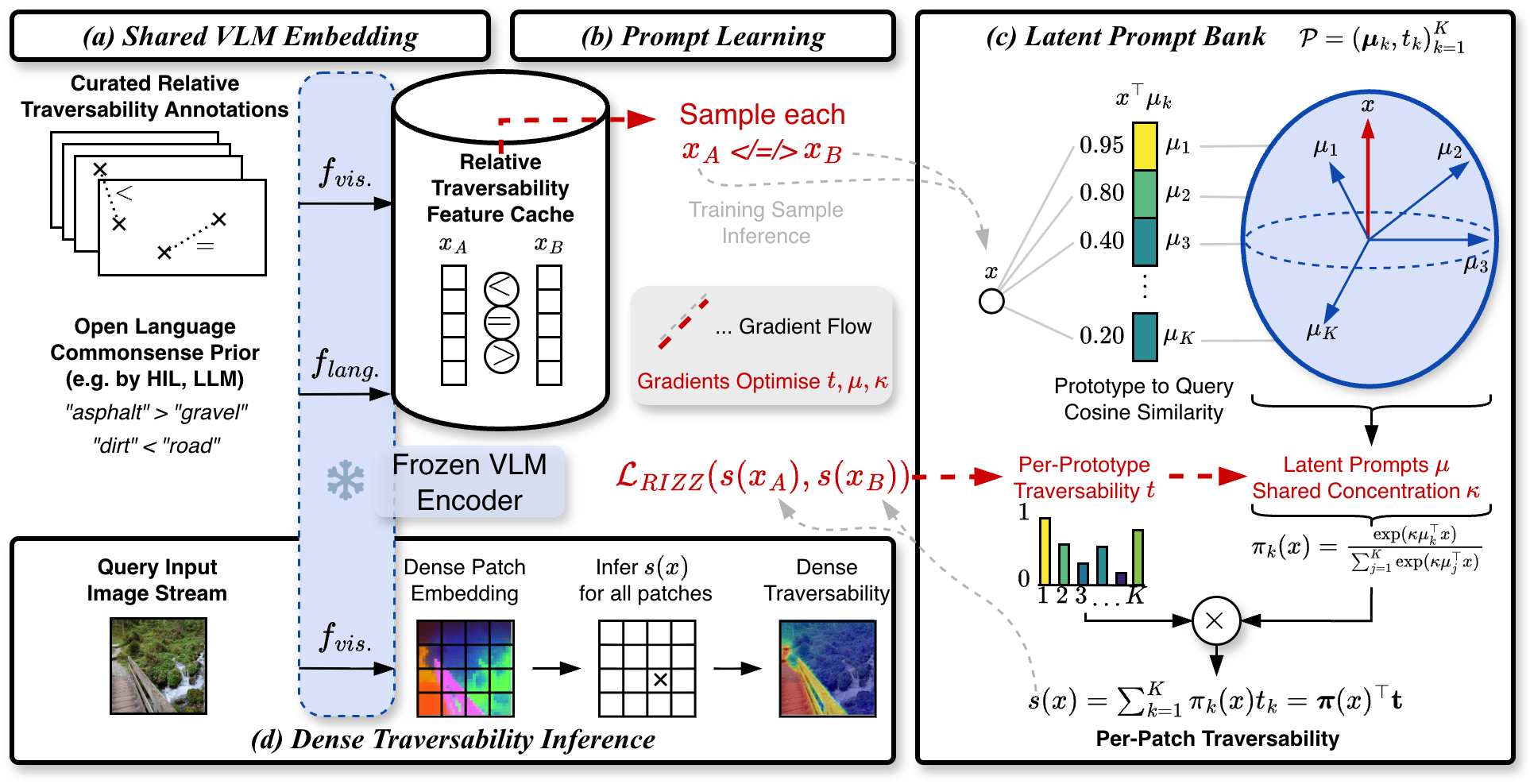}
    \vspace{-2mm}
    \caption{\textbf{Method Overview.} (a) relative traversability annotations in language and image modalities are encoded using a frozen VLM. (b) encoded annotation pairs are cached. A square hinge loss is applied over all cached annotations to parametrise a vMF mixture learning latent prototypes in VLM feature space using gradient descent. (c) The estimator builds a traversability estimate using learned per-prototype traversability $t$, prototype directions $\mu$ and shared concentration $\kappa$. (d) Dense traversability is estimated for each encoded image patch, or in the case of language-embedded pointclouds, for each point.}
    \label{fig:method_summary}
    \vspace{-2mm}
\end{figure*}

This paper introduces a method for learning and adapting language priors using sample-efficient traversability preference learning.
Motivated by recent successes of representing semantic classes as mixtures in language-grounded feature spaces \cite{schwaiger2026resireg, maggio2026foundit}, traversability is modelled as latent prototypes to leverage the already abstract VLM representation without application head training. 
Prototypes lie in the vision-language embedding space and serve as VLM prompts for downstream TE queries.
Fig.~\ref{fig:method_summary} summarises our method. 
First, a frozen VLM encodes relative traversability pairs from image annotations and open language vocabulary in the shared feature space (a).
A relative ranking loss is applied over each pair to parametrise the mixture's per-prototype traversability $t$, per-prototype direction in latent space $\mu$, and shared concentration $\kappa$ (b).
Traversability preference is modelled using vMF mixture responsibilities, with per-prototype traversability weighted by cosine similarity between query feature and per-prototype direction $\mu$ (c).
We apply the trained estimator to each feature in a query feature map to infer dense traversability (d).

\subsection{Shared Vision-Language Embedding}
Our method processes relative supervision signals, consisting of tuples $(x_A,x_B,r)$ with $r\in\{-1,0,+1\}$; $r=+1$ denotes that $B$ is more traversable than $A$, and vice-versa.
A frozen VLM encoder adjacently embeds images and natural language using two encoders: $f_{vis.}$ for images and $f_{lang.}$ for natural language prompts.
The visual encoder $f_{vis.}$ maps input images $I \in \mathbb{R}^{W \times H \times 3}$ to a lower resolution feature map $f_{vis.}: I \to \mathbb{R}^{W/16 \times H/16 \times D}$, with $D$ depicting embedding dimension.
Since relative image annotations require embeddings at specific pixel indices in the image, the feature map is bilinearly upsampled to pixel-level and sampled at the correct position.
The language encoder $f_{lang.}$ embeds natural language prompts into a single feature vector of embedding dimension $D$.

\subsection{Prompt Learning}
All embedded relative annotation pairs are gathered for downstream estimator training in the form $x_A, x_B \in \mathbb{R}^D$.
Forward inference yields estimated absolute traversability $s(x)$ from a single embedding $x$.
Endpoint scores $s_A=s(x_A)$ and $s_B=s(x_B)$ yield the relative difference $\Delta=s_B-s_A$ used for training on relative annotations.
Following \cite{schreiber2024wrizz}, a squared hinge loss and squared difference are applied to inequality and equality annotations respectively (see Eq.~\ref{eq:lrizz_loss}).
$N_{\neq0}$ and $N_{=0}$ denote the number of non-tie and tie comparisons respectively and $r$ denotes the relative annotation. 
\begin{equation}
    \mathcal{L}_{\mathrm{RIZZ}}=
    \frac{1}{N_{\neq 0}}\sum_{r\neq 0}\left[\max(0,L-r\Delta)\right]^2+
    \frac{1}{N_{=0}}\sum_{r=0}\Delta^2
    \label{eq:lrizz_loss}
\end{equation}

\textbf{Language Initialisation.}
CLIP-style VLMs rely on the final linear projection head to align both encoders \cite{zhou2022maskclip}.
While this effectively increases absolute embedding similarity between related semantic concepts across encoders, a modality gap remains w.r.t. identifiability and embedding scale. 
We therefore use modality-specific training configurations.
During initial training on open-language, the prototype directions are set to the spherical $k$-means centroids of $\{e_i\}$, i.e., the maximum-likelihood mean directions of a shared-$\kappa$ vMF mixture fitted to the vocabulary, and are held fixed while $\mathcal{L}_{\mathrm{RIZZ}}$ fits only the utilities $\mathbf{t}$ and the concentration $\kappa$ \cite{banerjee2005clustering}.

Fixing $\mu$ under language supervision follows from an identifiability argument. 
The vocabulary has a near-bipartite comparison graph and does not constrain $K$ directions in $\mathbb{S}^{D-1}$.
Therefore, free prototypes may be reduced into a two-prototype linear separator that carries no information about image embeddings due to the modality gap.
Anchoring the directions to the concept centroids instead preserves the mechanism behind zero-shot transfer in CLIP-style models: an image patch is assigned to prototypes by its similarity to language concepts, and the language comparisons decide only how those concepts are valued \cite{radford2021learning}.

\textbf{Concentration Transfer.}
The VLM modality gap also affects the scale of the prompt bank.
Concept-prototype cosines are larger and more widely spread than patch-prototype cosines, so a concentration $\kappa_{\mathrm{text}}$ fitted on concepts yields flatter responsibilities on image patches, and $s(x)$ contracts toward the mean utility.
Because $\kappa$ only sets assignment sharpness, it can be transferred without labels: We choose $\kappa_{\mathrm{img}}$ such that the mean responsibility entropy of unlabeled deployment-domain patches matches that of the language concepts (see Eq.~\ref{eq:concentration_transfer}).
\begin{equation}
    \mathbb{E}_{x}
    H\!\left(\pi(x;\kappa_{\mathrm{img}})\right)
    =
    \mathbb{E}_{i}
    H\!\left(\pi(e_i;\kappa_{\mathrm{text}})\right)
    \label{eq:concentration_transfer}
\end{equation}
$x$ ranges over patches from unannotated frames, $i$ uniformly indexes the language concepts, and $H(\boldsymbol{\pi})= -\sum_{k=1}^{K}\pi_k\log\pi_k$.
For non-degenerate prototype similarities, the mean patch entropy decreases monotonically with $\kappa$, allowing $\kappa_{\mathrm{img}}$ to be found by bisection.
Eq.~\ref{eq:concentration_transfer} is applied after language prior training to enable subsequent image-based fine-tuning, where the language-initialised bank is the starting point and $\mu$ is unfrozen.
Image-based fine-tuning and downstream TE operate in the same modality.
Without the modality transfer, the prototype directions can adapt to traversability preferences.
In practice, the optimiser updates unconstrained prototype vectors with subsequent normalisation during each forward pass.

\subsection{Latent Prompt Bank}
Instead of training a prediction head on top of the frozen backbone, we propose to model traversability using a prompt bank $\mathcal{P}=\{(\mu_k,t_k)\}_{k=1}^{K}$ containing normalised visual prototypes $\mu_k\in\mathbb{S}^{D-1}$ and bounded traversability utilities $t_k=\sigma(\tilde{t}_k)\in(0,1)$.
An advantage of this approach over an application head is the ability to explicitly handle modality mismatches in the encoder, enabling estimator bootstrapping from the language prior.
A learned shared concentration $\kappa=\mathrm{softplus}(\tilde{\kappa})+\epsilon$ controls the sharpness of prototype assignments.
Shared-concentration vMF responsibilities $\pi_k$ softly assign query feature $x$ to all prototypes weighted by cosine similarity (see Eq.~\ref{eq:vMF_responsibilities}).
\begin{equation}
    \pi_k(x)=
    \frac{\exp\!\left(\kappa\mu_k^\top x\right)}
    {\sum_{j=1}^{K}\exp\!\left(\kappa\mu_j^\top x\right)}
    \label{eq:vMF_responsibilities}
\end{equation}
Per-patch traversability $s(x)$ is retrieved from the responsibility-weighted prototype utility, combining the affinity $\pi_k$ between query and each prototype with the learned per-prototype traversability $t$ (see Eq.~\ref{eq:traversability_mixture}).
\begin{equation}
    s(x)=\sum_{k=1}^{K}\pi_k(x)t_k
    =\pi(x)^\top t
    \label{eq:traversability_mixture}
\end{equation}

\subsection{Dense Traversability Inference}
The resulting estimator predicts an absolute traversability value $\in [0,1]$ from a single VLM embedding.
For dense inference, query images are embedded using $f_{vis.}$.
The estimator is then applied to each image patch, with the resulting traversability map being bilinearly upsampled to pixel level.
This inference mode behaves similarly to existing image to image estimators predicting $I \in \mathbb{R}^{H \times W \times 3} \to T \in \mathbb{R}^{H \times W}$ and can be queried during in the field fine-tuning.
New annotations (e.g., obtained through human-in-the-loop (HIL) labelling, LVLM queries, or system-level self-labelling) are embedded and cached during operation, allowing for continual estimator parameter updates.

Our estimator operating from single VLM feature vectors offers an additional advantage for deployment.
Our method is not limited to end-to-end whole-image-processing.
Instead, it can also operate on semantic maps originating from VLM mapping methods.
In this case, instead of applying to each patch feature, the estimator is applied to each VLM embedding at a map's query points.

\section{Experiments}

\textbf{Evaluation Protocol and Metrics.}
Following \cite{schreiber2024wrizz}, we quantitatively evaluate Human Disagreement Rate (HDR) on a modified version of the WayFAST dataset \cite{gasparino2022wayfast}.
The modified WayFAST dataset consists of relative traversability annotations within and across images.
It has $6,805$ inequality and $13,067$ equality labels across $13,248$ images in the train set and $1,738$ inequality and $3,227$ equality labels across $3,310$ images in the validation set.
HDR denotes the fraction of predicted relations that disagree with the human-provided labels (see Eq.~\ref{eq:hdr_eval}).
\begin{equation}
    \mathrm{HDR}_\tau =
    \frac{1}{N}\sum_{i=1}^{N}
    \mathbf{1}\!\left[L_\tau(p_{i,a},p_{i,b})\neq t_i\right]
    \label{eq:hdr_eval}
\end{equation}
$N$ is the number of annotated pairs, $p_{i,a}$, $p_{i,b}$ are their predicted scores, $t_i$ is the human-provided relation, $L_\tau$ is the thresholded predicted relation, and $\mathbf{1}[\cdot]$ denotes the indicator function.
Following \cite{schreiber2024wrizz}, HDR is reported as a percentage for $\tau\in\{0.1,0.25,0.5\}$, where $\tau$ defines the maximum score difference considered equally traversable.
Lower score at higher thresholds is better, indicating clearer distinction between traversability classes.
To isolate prediction head performance, all VLM predictors use the same frozen encoder.

Quantitative results address three applications: Zero-shot, dataset transfer, and fine-tuned.
Zero-shot applicability reports the performance of traversability estimators not trained or tuned on the target domain. 
This quantifies our language prior solely trained on generic commonsense traversability vocabulary.
Following \cite{schreiber2025do}, we additionally report training-free dataset transfer from RUGD \cite{wigness2019RUGD} to WayFAST, depicting the transfer from road-based outdoor navigation to mostly off-road scenes.
Finally, we report fine-tuned results on the WayFAST training set.
To ablate the language prior, we report our estimator with random initialisation as well as language prior initialisation. 

To demonstrate close to real-world applicability, we additionally report qualitative results across the large-scale RoboNav dataset \cite{eder2023traversability}.
Qualitative results show TE on image-inputs as well as language-embedded semantic maps.

\begin{figure*}[ht]
    \vspace{2mm} 
    \centering
    \includegraphics[width=0.99\linewidth]{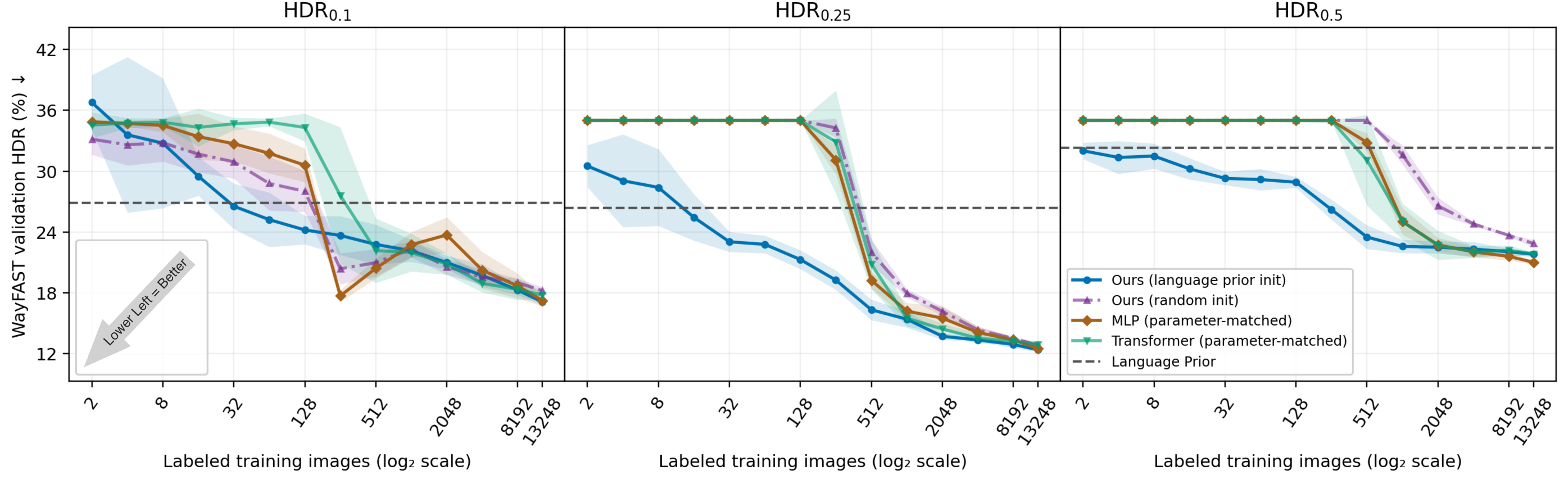}
    \vspace{-2mm} 
    \caption{\textbf{Fine-Tuning Sample efficiency on WayFAST} for language-prior and randomly initialised vMF prototypes, a parameter-matched MLP, as well as a parameter-matched transformer head, each with $\approx8.2$k trainable parameters. 
    All configurations are trained over $5$ seeds on identical nested image subsets, plotting mean over the results.
    Shading indicates $95$\% confidence interval.
    Dashed lines indicate the language prior.
    Our language-prior initialised estimator (blue line) is lower left compared to the other models, indicating improved fine-tuning sample-efficiency.}
    \label{fig:epx2_sample_efficiency}
    \vspace{-1mm}
\end{figure*}

\textbf{Baselines.}
Baseline choice is complex due to the close relationship between traversability estimation and the downstream planning in system-level TE research.
Training and evaluating solely on relative labels is advantageous, since other information sources can easily be converted to relative annotations.
Examples are self-labelling from robot experience, HIL annotations, and in-the-field generated LVLM predictions.
However, this limits the applicability of self-labelling baselines to the experimental setup.

The baseline for zero-shot evaluation is a statically prompted VLM, following prior demonstrations of language representations' ability to represent fuzzy classes relevant to outdoor navigation \cite{schwaiger2025otas}.
It estimates traversability as temperature-scaled softmax over normalised cosine similarities.
Dataset transfer is evaluated against a HIL-focused baseline \cite{schreiber2025do} to show the accuracy-to-sample-efficiency trade-off of foundation model-based prediction heads.
Fine-tuned evaluation is conducted against the SOTA of end-to-end trained TE from relative annotations \cite{schreiber2024wrizz}.
To show the methodological advantage of our formulation against classic application head training, a linear probe, parameter-matched MLP, and parameter-matched transformer head on top of the same VLM are evaluated. 

\textbf{Implementation Detail.}
All VLM-based configurations use the ReSiReg Mini VLM \cite{schwaiger2026resireg} due to its computationally efficient, low-parameter (25M) visual encoder with feature reconstruction disabled to prevent possible oversmoothing in the vision-language feature space.
Backbone, prediction heads and our estimator are implemented using PyTorch \cite{pytorch}.
All experiments use $K=16$ prototypes unless stated differently.
Image-based training lasts $30$ epochs of $2,000$ iterations (regardless of training sample count) using Adam \cite{kingma2015adam} with a learning rate of $\alpha=0.01$, batches of $64$ image pairs, and a margin of $\tau=0.5$ for the loss (see Eq.~\ref{eq:lrizz_loss}).
We maintain exponential moving average (EMA) of estimator parameters while training with image supervision using decay factor $\beta=0.99$.
All evaluations use the same full-precision compute to prevent potential numerical instabilities of baselines on an Nvidia RTX6000 GPU using fixed seeds. 

\textbf{Language Prior Pretraining.}
To improve downstream fine-tuning sample efficiency and TE accuracy, configurations denoted as \textit{(language prior init)} are initialised from a generic open language vocabulary of relative traversability rules.
During image-based fine-tuning, the language-based traversability rules are aligned with the target domain and its preferences.
To ensure no domain-specific contamination of the language prior, the training vocabulary is generated using an LLM (GLM-5.3-Flash). 
It consists of $300$ inequality samples and $7$ equality samples.
The language prior is trained for $30$ epochs consisting of $2000$ iterations uniformly sampled from the training vocabulary using Adam with learning rate $\alpha=0.01$. 
This vocabulary is also used to prompt the statically prompted baseline in experiment~1.
To provide a best-case scenario for the baseline and ensure a fair comparison that does not favour our method, the single prompt \textit{"tarmac"}, reporting the best downstream performance is used.

\subsection{Human Disagreement Traversability Estimation} \label{sec:experiment1}
Tab.~\ref{tab:hdr_relative_traversability} quantifies overall accuracy across estimators.
For zero-shot application (Tab.~\ref{tab:hdr_relative_traversability}a), the performance delta between the statically prompted VLM and language prior highly depends on the HDR threshold $\tau$, where the static prompt wins HDR$_{0.1}$ and the language prior wins the other thresholds. 
Tab.~\ref{tab:hdr_relative_traversability}b) reports dataset transfer by training on a custom dataset following \cite{schreiber2025do} assembled from RUGD annotations and evaluating on WayFAST.
The baseline transfers better across datasets, possibly due to the increased application head size and the more abstract self-supervised learning (SSL) representation used as the baseline's backbone.
The baseline's trained traversability estimator head size is $\approx49.4$k parameters (disregarding the uncertainty estimator), while ours is significantly smaller at $\approx8.2$k.
While SSL features have been shown superior for spatial tasks \cite{man2024lexicon3d}, the enhanced spatial structure of SSL features contradicts view-dependent noise in language-grounded backbones \cite{schwaiger2026resireg}.
Interestingly, when comparing vMF initialisation methods for this training-free transfer across datasets, randomly initialising our prototypes yields slightly improved HDR$_{0.1}$ and HDR$_{0.25}$.
The language prior improves results on HDR$_{0.5}$, surpassing the baseline.
This indicates that language prior initialisation pushes per prototype traversability further towards the bounds of $0$ and $1$.

Tab.~\ref{tab:hdr_relative_traversability}c) presents fine-tuned results trained on the WayFAST train set.
Linearly probing the VLM does not yield similar accuracy to other fine-tuned configurations.
Ours with random initialisation yields a measurable improvement.
\cite{schreiber2024wrizz} is an end-to-end model trained on this dataset using the same loss as our predictors and serves as the SOTA for this benchmark.
Ours with language prior initialisation wins the two higher HDR thresholds, where the metric demands clearer distinction between semantic traversability modes.

\begin{table}[t]
    \centering
    \caption{\textbf{WayFAST Traversability Estimation}. Our method for estimating traversability in VLM feature space is competitive with purpose-trained end-to-end estimator networks, while minimising the amount of trained parameters.
    Results are highlighted as\\
    \protect\colorbox{green!25}{\strut\textbf{best}},
    \protect\colorbox{yellow!30}{\strut second-best}, and
    \protect\colorbox{orange!30}{\strut third-best}.
    }
    \begin{tabular}{l|c|c|c c}
        \hline
        \textbf{(a) Zero-Shot} & HDR$_{0.1}$ & HDR$_{0.25}$ & HDR$_{0.5}$ & (\%)$\downarrow$ \\
        VLM + Static Prompt & 23.9 & 34.4 & 35 \\
        Language Prior (Ours) & 27.0 & 26.3 & 32.3 \\
        \hline
        \textbf{(b) Dataset Transfer} & HDR$_{0.1}$ & HDR$_{0.25}$ & HDR$_{0.5}$ & (\%)$\downarrow$ \\
        Big Chungus \cite{schreiber2025do} & 19.9 & 16 & 27.4 \\
        Ours (random init) & 29.1 & 19.0 & 27.8 \\
        Ours (language prior init)& 30.9 & 19.9 & 24.11 \\
        \hline
        \textbf{(c) Fine-Tuned} & HDR$_{0.1}$ & HDR$_{0.25}$ & HDR$_{0.5}$ & (\%)$\downarrow$ \\
        W-RIZZ \cite{schreiber2024wrizz} & \firstbest{16.5} & \secondbest{12.5} & \secondbest{22.5} \\
        VLM + Linear & 24.6 & 16.4 & 23.7\\
        Ours (random init) & \thirdbest{18.4} & \thirdbest{12.8} & \thirdbest{22.8}\\
        Ours (language prior init)& \secondbest{17.1} & \firstbest{12.3} & \firstbest{21.8}\\
        \hline
    \end{tabular}
    \label{tab:hdr_relative_traversability}
    \vspace{-2mm}
\end{table}


\begin{figure*}[ht]
    \vspace{2mm} 
    \centering
    \includegraphics[width=0.96\linewidth]{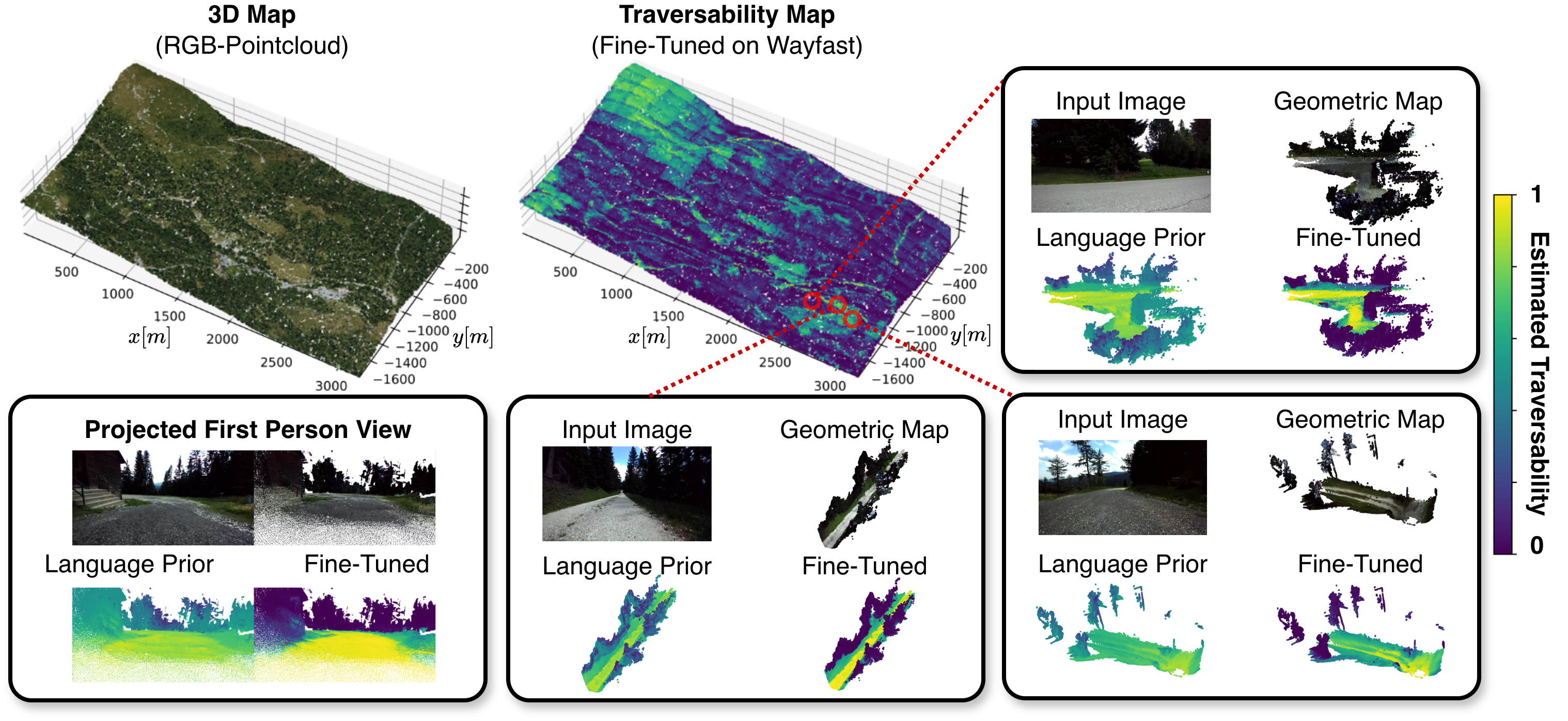}
    \caption{\textbf{Qualitative Traversability Estimation on RoboNav.} Traversability predictions for outdoor scenes from the RoboNav Husky Route 1 recording. We compare the language prior to the language prior-initialised prototype bank adapted on the complete WayFAST training split. Predictions are shown for both image features and reconstructed 3D semantic maps using the same fixed $[0,1]$ scale, where higher values indicate greater estimated traversability.}
    \label{fig:exp3_traversability_map}
    \vspace{-2mm}
\end{figure*}

\subsection{Language-Guided Adaptation Across Annotation Budgets}
Since field experiments typically provide only limited amounts of training data, we additionally evaluate whether language-guided initialisation improves sample efficiency during online adaptation.
Language prior initialised and randomly initialised vMF mixtures are evaluated against parameter-matched MLP and parameter-matched transformer application heads.
The MLP uses a $512 \rightarrow 16 \rightarrow 1$ architecture with ReLU hidden activations and a sigmoid output. 
The transformer partitions the single $512$-dimensional input feature into $16$ consecutive groups with learned embedding and prepended learned CLS token.
Tokens use a shared linear projection and are passed through a pre-norm (width $16$) $\to$ $2\times$attention head architecture with subsequent feedforward of $16 \to 194 \to 16$ with GELU activation.
Scalar TE is output as a linear sigmoid readout.
The heads have been chosen to ensure a fair comparison w.r.t. trainable head capacity, with both heads resulting in $\approx8.2$k trainable parameters, similar to our vMF mixture at $K=16$. 
To measure sample efficiency, we construct nested training subsets containing $2,4,8,\ldots,8192$ images, together with the complete $13,248$-image training split.
Sampling preserves the dataset's disjoint image pairs, with each pair providing three relative comparisons.
An $N$-image budget therefore corresponds to $3N/2$ annotations.
All methods receive identical subsets and minibatch orders.
For each budget, the language-initialised vMF mixture starts from the same text-derived prior, while the random prototype and MLP baselines are randomly initialised.
We train each configuration across $5$ seeds and report mean resulting HDR.
Shading indicates $95$\% confidence interval over WayFAST validation HDR.

Fig.~\ref{fig:epx2_sample_efficiency} presents results across HDR thresholds.
All configurations without language prior exhibit similar sample-efficiency.
This is expected given similar model capacities.
HDR$_{0.1}$ results show larger spread across runs, since the low threshold is less strict w.r.t. difference in absolute traversability predictions.
HDR$_{0.25}$ and HDR$_{0.5}$ show clear separation between our language-prior initialised vMF mixture and all other configurations.
There, our approach improves downstream accuracy at low training sample sizes.

\subsection{3D Semantic Map Queries}
Since $s(x)$ predicts traversability directly from query VLM feature $x$, TE is not bound to full image queries.
Instead, our estimator can perform TE on language-embedded 3D maps obtained through recent VLM semantic mapping methods \cite{schwaiger2025otas}.
We qualitatively demonstrate this without additional training or scene-specific adaptation using the language-initialised estimator trained in Sec.~\ref{sec:experiment1}.
Fig.~\ref{fig:exp3_traversability_map} presents a large scale language-embedded map assembled from RoboNav data \cite{eder2023traversability} with our traversability estimator applied.
The map is a RGB pointcloud, where each point has an associated VLM embedding. 

The large aerial RGB and traversability maps are assembled from the RoboNav orthophoto and corresponding elevation raster.
For semantic mapping, the orthophoto is split into $240 \times 424$-pixel tiles with each tile being resized to $480 \times 848$ pixels for VLM inference.
The resulting feature grid with $4$ metre spacing is bilinearly interpolated to $1$ metre resolution with subsequent grid-wise TE.

Additionally, the small map excerpts use 3D projection from camera intrinsics and extrinsics estimated using VGGT~\cite{wang2025vggt} on robot data from the \textit{Husky Route 1} recording.
We provide TE on this language-embedded semantic map using the language-prior and the estimator fine-tuned on WayFAST.
Language prior predictions qualitatively observe less clear distinction between most and least traversable areas in the scene.
Meanwhile, the fine-tuned estimator pushes predictions of untraversable obstacles (e.g., trees) towards $0$.
The fine-tuned estimator also properly encapsulates the gradient from the fully traversable street (yellow) over partly traversable grass (yellow) to untraversable obstacles (dark purple).

\subsection{Ablations}

\begin{figure*}[ht]
    \vspace{2mm} 
    \centering
    \includegraphics[width=0.9\linewidth]{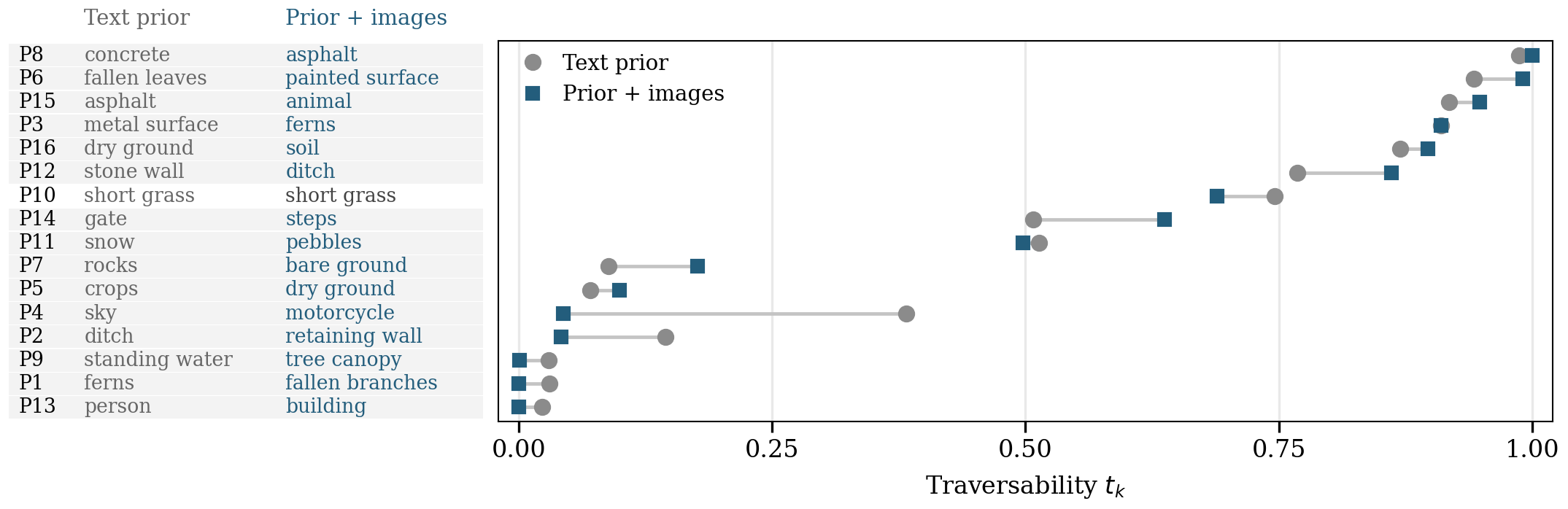}
    \vspace{-2.5mm} 
    \caption{\textbf{Prototype semantics and traversability scores} before and after image-based fine-tuning. Left: CLIP-Dissect descriptions inferred for each prototype direction $\mu$ from the language-prior and the fine-tuned estimator. Right: corresponding traversability scores $t_k$, with gray circles denoting the text prior and blue squares the image-fine-tuned prototypes.}
    \label{fig:prototype_semantics}
    \vspace{-2mm}
\end{figure*}

\textbf{Prototype Sensitivity.}
Tab.~\ref{tab:k-sweep} ablates downstream accuracy using a sweep over $K$ for fine-tuned and dataset transfer evaluations on WayFAST (random init).
Results show low overall sensitivity to $K$ with differing top scores across HDR thresholds.
However, depending on training and target domain, the larger number of training parameters from higher $K$ may be beneficial; indicated by the top scores on Tab.~\ref{tab:k-sweep}b) arising at $K=100$.
We chose $K=16$ since it offers a good tradeoff between capacity and processing speed with diminishing returns from additional prototypes.

\begin{table}[t]
    \centering
    \caption{vMF prototype-capacity ablation on WayFAST validation set. Results show HDR (\%); lower is better.}
    \begin{tabular}{l|c|c|c c}
        \hline
        \textbf{(a) Dataset Transfer} & HDR$_{0.1}$ & HDR$_{0.25}$ & HDR$_{0.5}$ & (\%)$\downarrow$ \\
        $K=4$   & 29.3 & 19.0 & 27.4 \\
        $K=8$   & \firstbest{27.8} & 18.3 & \firstbest{27.2} \\
        $K=12$  & 30.1 & 19.6 & 27.6 \\
        $K=16$  & 29.1 & 19.0 & 27.8 \\
        $K=24$  & 29.6 & 19.3 & 27.5 \\
        $K=32$  & 29.1 & 19.0 & \firstbest{27.2} \\
        $K=40$  & 28.4 & 18.4 & 28.0 \\
        $K=48$  & 28.4 & 18.3 & 27.3 \\
        $K=64$  & 28.6 & \firstbest{17.8} & 27.7 \\
        $K=80$  & 28.2 & 18.0 & 27.8 \\
        $K=100$ & 29.4 & 18.9 & 27.9 \\
        \hline
        \textbf{(b) Fine-Tuned} & HDR$_{0.1}$ & HDR$_{0.25}$ & HDR$_{0.5}$ & (\%) $\downarrow$ \\
        $K=4$   & 18.6 & 13.7 & 23.5 \\
        $K=8$   & 18.9 & 13.3 & 22.9 \\
        $K=12$  & 18.9 & 13.3 & 23.0 \\
        $K=16$  & 18.4 & 12.8 & 22.8 \\
        $K=24$  & 18.5 & 12.6 & 22.9 \\
        $K=32$  & 18.3 & \firstbest{12.4} & 22.6 \\
        $K=40$  & 18.3 & 12.6 & 22.6 \\
        $K=48$  & 18.0 & 12.7 & 22.7 \\
        $K=64$  & 18.3 & 12.7 & 22.5 \\
        $K=80$  & \firstbest{17.7} & 12.6 & 22.6 \\
        $K=100$ & 18.4 & \firstbest{12.4} & \firstbest{22.4} \\
        \hline
    \end{tabular}
    \label{tab:k-sweep}
\end{table}

\textbf{Language Prior Training Vocabulary Size.}
Tab.~\ref{tab:zs-vocab-sweep} evaluates the effect of language-supervision scale at a fixed prototype count of $K=16$.
We construct two language-supervision datasets.
(1) a \emph{tiny} dataset containing $73$ unique descriptions and 307 pairwise comparisons.
(2) a larger nested dataset derived from a vocabulary of 363 descriptions. From this vocabulary set, we construct subsets of $50$, $100$, $200$, and $363$ unique descriptions and retain all pairwise comparisons whose endpoints are contained in the respective subset.
The \emph{tiny} vocabulary achieves the lowest HDR at all three thresholds despite containing only 307 comparisons.
HDR$_{0.1}$ decreases as the vocabulary grows from \emph{small} to \emph{large}, but slightly increases again for \emph{extra large}.
In contrast, HDR$_{0.25}$ varies irregularly and HDR$_{0.5}$ remains effectively unchanged across all nested variants.
Thus, instead of increasing the total number of language comparisons, the number, composition and balance of generated descriptions seem to be important to downstream accuracy.
This could be leveraged by tailoring the language prior to an expected use case, e.g., by manually tuning the vocabulary or specifying in the prompt.

\begin{table}[t]
    \centering
    \caption{Language-prior vocabulary size ablation on WayFAST validation set. Results show HDR (\%); lower is better.}
    \begin{tabular}{l|c|c|c|c c}
        \hline
        \textbf{Vocabulary} & \#Pairs & HDR$_{0.1}$ & HDR$_{0.25}$ & HDR$_{0.5}$ & (\%)$\downarrow$ \\
        tiny & 307 
            & \firstbest{27.0} & \firstbest{26.3} & \firstbest{32.3} \\
        small & 521 
            & 31.1 & 30.5 & 34.8 \\
        medium & 2,073 
            & 29.0 & 30.3 & 34.9 \\
        large & 8,337 
            & 28.0 & 32.4 & 34.9 \\
        extra large & 27,281 
            & 28.4 & 29.3 & 34.9 \\
        \hline
    \end{tabular}
    \label{tab:zs-vocab-sweep}
\end{table}

\textbf{Prototype to Semantic Class Matching.}
To test whether prototype scores have an interpretable semantic meaning, we compare the text-prior and image-fine-tuned prototype banks using descriptions inferred with CLIP-Dissect~\cite{oikarinen2023clipdissect}, adapted to our used VLM.
Fig.~\ref{fig:prototype_semantics} presents the results.
Several prototypes show plausible associations between their inferred semantics and traversability scores.
In particular, the image-fine-tuned bank assigns high scores to prototypes described as \emph{asphalt} and \emph{soil}, while prototypes described as \emph{retaining wall} and \emph{building} receive low scores.
At the same time, the inferred descriptions also show cases where the semantic interpretation is less consistent.
For example, \emph{ditch} receives a high score, and prototype 4 changes from \emph{sky} in the text-prior bank to \emph{motorcycle} after image fine-tuning.
One possible explanation for such changes is ambiguity in the frozen language-embedding space.
For example, \emph{car} is the sixth-nearest candidate to \emph{sky}, with a cosine similarity of $0.59$ between their embeddings.
The used VLM was chosen to minimise computational requirements for fine-tuning onboard of autonomous mobile robots.
The therefore lower parameter count and limited representational capacity might further contribute to this ambiguity.

\section{Conclusion}

This paper introduced a traversability estimator that represents terrain preferences as language-initialised vMF prototypes in a frozen vision-language feature space.
The proposed method enables bootstrapping domain-specific fine-tuning of traversability preferences using a natural language prior.
Quantitative experiments show that language initialisation improves sample efficiency and that the fine-tuned estimator achieves competitive accuracy against the SOTA.
Qualitative results further demonstrate the estimator's applicability to language-embedded 3D semantic maps and examine semantic interpretations of the learned prototypes.
Future work will investigate continual learning from robot experience, applicability to general-purpose affinity learning, and improved interpretation of learned prototype semantics.

{\small
\bibliographystyle{IEEEtran}
\bibliography{ref}
}

\end{document}